\documentclass[sigconf]{acmart}

\AtBeginDocument{%
  }

\setcopyright{acmlicensed}
\copyrightyear{2018}
\acmYear{2018}
\acmDOI{XXXXXXX.XXXXXXX}
\acmConference[Conference acronym 'XX]{Make sure to enter the correct
  conference title from your rights confirmation email}{June 03--05,
  2018}{Woodstock, NY}
\acmISBN{978-1-4503-XXXX-X/2018/06}

\usepackage[utf8]{inputenc}
\usepackage[most]{tcolorbox}
\tcbuselibrary{listings} 
\usepackage{geometry}
\usepackage{multicol} 
\usepackage{multirow}

\usepackage{microtype}

\usepackage{url}
\usepackage{graphicx} 
\usepackage{float} 
\usepackage{booktabs} 
\usepackage{amsmath}

\usepackage{array}
\usepackage{tabularx}
\usepackage{booktabs}
\usepackage[normalem]{ulem}
\useunder{\uline}{\ul}{}
\usepackage{booktabs} 
\usepackage{enumitem}

\usepackage{caption}
\usepackage{xcolor} 
\definecolor{mygreen}{RGB}{34, 139, 34}

\newtcolorbox[list inside=prompt,auto counter]{prompt}[1][]{
    colbacktitle=black!60,
    coltitle=white,
    fontupper=\normalsize,
    boxsep=5pt,
    left=0pt,
    right=0pt,
    top=0pt,
    bottom=0pt,
    boxrule=1pt,
    #1,
}

\begin{document}

\title{Learning to Shop Like Humans: A Review-driven Retrieval-Augmented Recommendation Framework with LLMs}

\author{Kaiwen Wei}
\affiliation{%
  \institution{Chongqing University}
  \city{Chongqing}
  \country{China}
}
\email{weikaiwen@cqu.edu.cn}

\author{Jinpeng Gao}
\affiliation{%
  \institution{Chongqing University}
  \city{Chongqing}
  \country{China}}
\email{gaojinpeng@stu.cqu.edu.cn}

\author{Jiang Zhong}
\authornotemark[1]
\affiliation{%
  \institution{Chongqing University}
  \city{Chongqing}
  \country{China}
}
\email{zhongjiang@cqu.edu.cn}

\author{Yuming Yang}
\affiliation{%
 \institution{Chongqing University}
 \city{Chongqing}
 \country{China}}
\email{ymyang@cqu.edu.cn}

\author{Fengmao Lv}
\affiliation{%
  \institution{Southwest Jiaotong University}
  \city{Chengdu}
  \country{China}}
\email{fengmaolv@swjtu.edu.cn}

\author{Zhenyang Li}
\authornote{Corresponding author}
\affiliation{%
  \institution{Hong Kong Generative AI Research and Development Center,}
  \institution{City University of Hong Kong}
  \city{Hong Kong}
  \country{China}}
\email{zhenyanglidz@gmail.com}

\renewcommand{\shortauthors}{Trovato et al.}

\begin{abstract}
Large language models (LLMs) have shown strong potential in recommendation tasks due to their strengths in language understanding, reasoning and knowledge integration. These capabilities are especially beneficial for review-based recommendation, which relies on semantically rich user-generated texts to reveal fine-grained user preferences and item attributes. However, effectively incorporating reviews into LLM-based recommendation remains challenging due to (1) inefficient to dynamically utilize user reviews under LLMs’ constrained context windows, and (2) lacking effective mechanisms to prioritize reviews most relevant to the user’s current decision context. To address these challenges, we propose RevBrowse, a review-driven recommendation framework inspired by the “browse-then-decide” decision process commonly observed in online user behavior. RevBrowse integrates user reviews into the LLM-based reranking process to enhance its ability to distinguish between candidate items. To improve the relevance and efficiency of review usage, we introduce PrefRAG, a retrieval-augmented module that disentangles user and item representations into structured forms and adaptively retrieves preference-relevant content conditioned on the target item. Extensive experiments on four Amazon review datasets demonstrate that RevBrowse achieves consistent and significant improvements over strong baselines, highlighting its generalizability and effectiveness in modeling dynamic user preferences. Furthermore, since the retrieval-augmented process is transparent, RevBrowse offers a certain level of interpretability by making visible which reviews influence the final recommendation.
\end{abstract}




\received{20 February 2007}
\received[revised]{12 March 2009}
\received[accepted]{5 June 2009}

\maketitle

\section{Introduction}

Recently, large language models (LLMs) have achieved remarkable success across a wide range of natural language processing tasks~\cite{DBLP:conf/acl/WeiSZZGJ20, DBLP:journals/pami/LiuWYTSYYJLF24, DBLP:conf/coling/WeiZZZZ0YZ25}, which demonstrate strong capabilities in understanding and generating human-like text~\cite{DBLP:journals/corr/abs-2402-06196, DBLP:journals/corr/abs-2501-09223}. Building on these advances, there has been growing interest in leveraging LLMs for recommender systems (RS). A number of recent studies~\cite{DBLP:journals/corr/abs-2405-05562, DBLP:conf/www/ZhengCQZ024, DBLP:journals/corr/abs-2304-10149} have explored how LLMs can be adapted to provide personalized, context-aware recommendations by formulating recommendation as a language modeling problem. This emerging paradigm has shown great potential in various aspects of recommendation, such as item ranking~\cite{DBLP:conf/www/BaeJMK24}, user preference modeling~\cite{DBLP:journals/tiis/JawaheerWK14}, and the interpretability of recommendation~\cite{explainable_recommendation}.


\begin{figure}[t]
\small
\centerline{\includegraphics[width=0.48\textwidth]{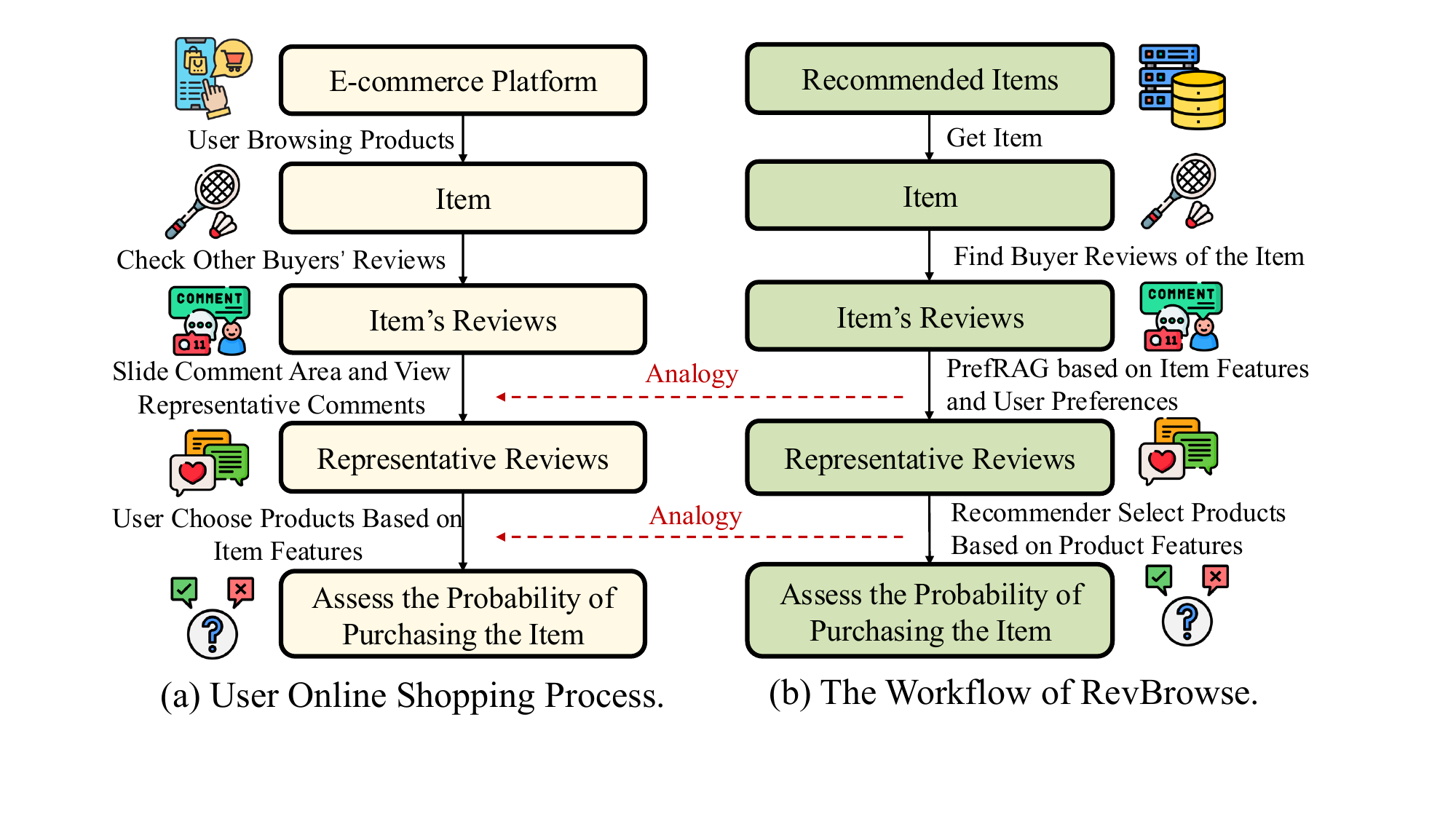}}
\caption{Analogy between the human shopping process and the proposed RevBrowse framework. RevBrowse emulates the way humans shop: its PrefRAG module represents the review-search phase, and its recommender corresponds to the final human decision-making step.}
\label{intro_sample}
\end{figure}

Among various approaches in RS, review-based recommender systems~\cite{DBLP:journals/eswa/WeiXZMP24, DBLP:conf/cikm/ZhangQLZZMC21} have shown particular promise by leveraging user-generated reviews to capture fine-grained preferences. Traditional approaches have employed  aspect modeling~\cite{DBLP:conf/recsys/McAuleyL13, DBLP:conf/ijcai/TanZLM16}, attention mechanisms~\cite{DBLP:journals/tkde/TalLHYA21}, and graph-based learning~\cite{DBLP:journals/kbs/LiuCLZW23} to improve recommendation accuracy and explainability. More recently, LLM-based methods have further advanced the field by incorporating prompt tuning~\cite{DBLP:journals/corr/abs-2306-01475}, in-context learning~\cite{DBLP:journals/tacl/RamLDMSLS23}, and generative reasoning~\cite{DBLP:journals/corr/abs-2408-06276} to extract semantic information from reviews and support more effective recommendations. Despite this progress, several key challenges remain:

A central challenge lies in the \textbf{limited efficiency of leveraging user reviews as preference signals}. Due to the constrained context window of LLMs, feeding many long reviews often leads to context truncation or excessive computational costs. Some studies attempt to mitigate this with review summarization~\cite{DBLP:journals/corr/abs-2408-06276}, where the summary is constructed from all available user reviews, regardless of recency. However, such fixed summaries often dilute or overlook recent changes in user preferences, failing to capture dynamically shifts in user preferences.


Building on this, another key challenge is the \textbf{lack of effective mechanisms to prioritize reviews that matter most}. In practice, only a small subset of reviews is truly pertinent to a user’s decision on a specific item. Without selective filtering, irrelevant or redundant content can overwhelm the model and obscure meaningful signals. Nevertheless, how to effectively identify and select the reviews that best capture human preferences remains a largely unexplored problem.




To better address these challenges, we turn to behavioral insights from real-world online shopping scenarios. As illustrated in Figure~\ref{intro_sample}~(a), users typically engage in a “browse-then-decide” process, during which they scan multiple reviews, each focusing on a different aspect such as comfort, design, or durability, before eventually forming a purchase decision. This observation motivates the design of a dynamic and adaptive review selection strategy that can more accurately reflect users’ evolving preferences in specific recommendation contexts.

Inspired by this human decision-making pattern, as illustrated in Figure~\ref{intro_sample}~(b), we propose RevBrowse, a review-centric recommendation framework that integrates retrieval and sequence modeling to enable LLMs to better capture long-term and dynamic user preferences. To emulate the experience of users browsing reviews before a purchase, we further introduce PrefRAG, a retrieval-augmented generation module that disentangles historical user preferences and item features into structured representations. PrefRAG utilizes contrastive learning to enhance the retrieval process, allowing the model to better retrieve more relevant features based on the target item. 
We conduct extensive experiments on the Amazon-2014 Food, Sports, Clothing, and Games datasets. Results show that RevBrowse significantly outperforms strong baselines on the sequential recommendation task. In addition, the transparent retrieval process in PrefRAG provides a degree of interpretability. 
In summary, the contributions of this paper are as follows: 

(1) Motivated by users’ “browsing-to-decision” behavior, we propose RevBrowse, a review-centric recommendation framework that leverages RAG strategy to efficiently compress lengthy review histories into a focused context window. It allows LLMs to better capture both long-term preferences and dynamically shifting user preferences.

(2) We design PrefRAG, a retrieval-augmented module that disentangles user and item features into structured representations, and dynamically selects relevant features conditioned on the target item, enabling effective and personalized preference matching.

(3) We conduct comprehensive experiments on four Amazon datasets, where RevBrowse consistently outperforms strong baselines and achieves state-of-the-art performance. Furthermore, the transparent retrieval process contributes to enhanced interpretability. All code will be released to support future research.









\section{Related Work}

\subsection{Review-based Recommendation}
Review-based recommender systems have attracted increasing attention due to their ability to leverage rich semantic cues in user reviews to enhance personalization and  accuracy. Early models relied on latent factor techniques and basic sentiment signals~\cite{DBLP:conf/recsys/McAuleyL13, DBLP:conf/ijcai/TanZLM16}, but recent advancements focus on extracting fine-grained preferences and item features through review-aware learning. Generic models like ConvMF~\cite{DBLP:journals/access/ChenFZWLSW19} learn user and item representations from raw review texts using convolutional neural networks (CNNs), while aspect-based approaches~\cite{DBLP:journals/tois/GuanCHZZPC19, DBLP:conf/ijcai/ChengD0ZSK18} capture user opinions on specific attributes using attention mechanisms~\cite{DBLP:journals/tkde/TalLHYA21}. Hybrid models~\cite{DBLP:journals/tnn/XiHWZL22} combining ratings and reviews have been shown to mitigate data sparsity and improve interpretability. More recently, graph-based methods~\cite{DBLP:journals/kbs/LiuCLZW23} utilize graph neural networks (GNNs) to model high-order user-item-aspect relations for better semantic reasoning. In parallel, contrastive learning~\cite{DBLP:journals/eswa/WeiXZMP24} has emerged as powerful paradigms for robust feature learning and dynamic user modeling. With the rise of Large Language Models (LLMs), models like PLLMPAR~\cite{DBLP:journals/corr/abs-2306-01475} and SIFN~\cite{DBLP:conf/cikm/ZhangQLZZMC21} integrate pre-trained transformers to understand reviews at a deeper semantic level, offering promising gains in both recommendation quality and transparency. Besides, Exp3rt~\cite{DBLP:journals/corr/abs-2408-06276} also proposed a step-by-step reasoning framework to enhance the rating prediction ability of LLMs. 

Despite their success, many review-based models encode user reviews into static representations, limiting their ability to adapt to specific recommendation contexts. This issue is exacerbated when applied to LLMs, as long review histories often exceed context window limits, leading to truncation and inefficiency. While some methods~\cite{DBLP:journals/corr/abs-2408-06276} use static summaries to compress inputs, they fail to reflect users’ dynamic, item-specific preferences. To address this, we propose a Retrieval-Augmented Generation (RAG) module named PrefRAG, which selectively retrieves the most informative features conditioned on the target item, enabling efficient and context-aware recommendation with LLMs.

\begin{figure*}[t]
\small
\centerline{\includegraphics[width=0.99\textwidth]{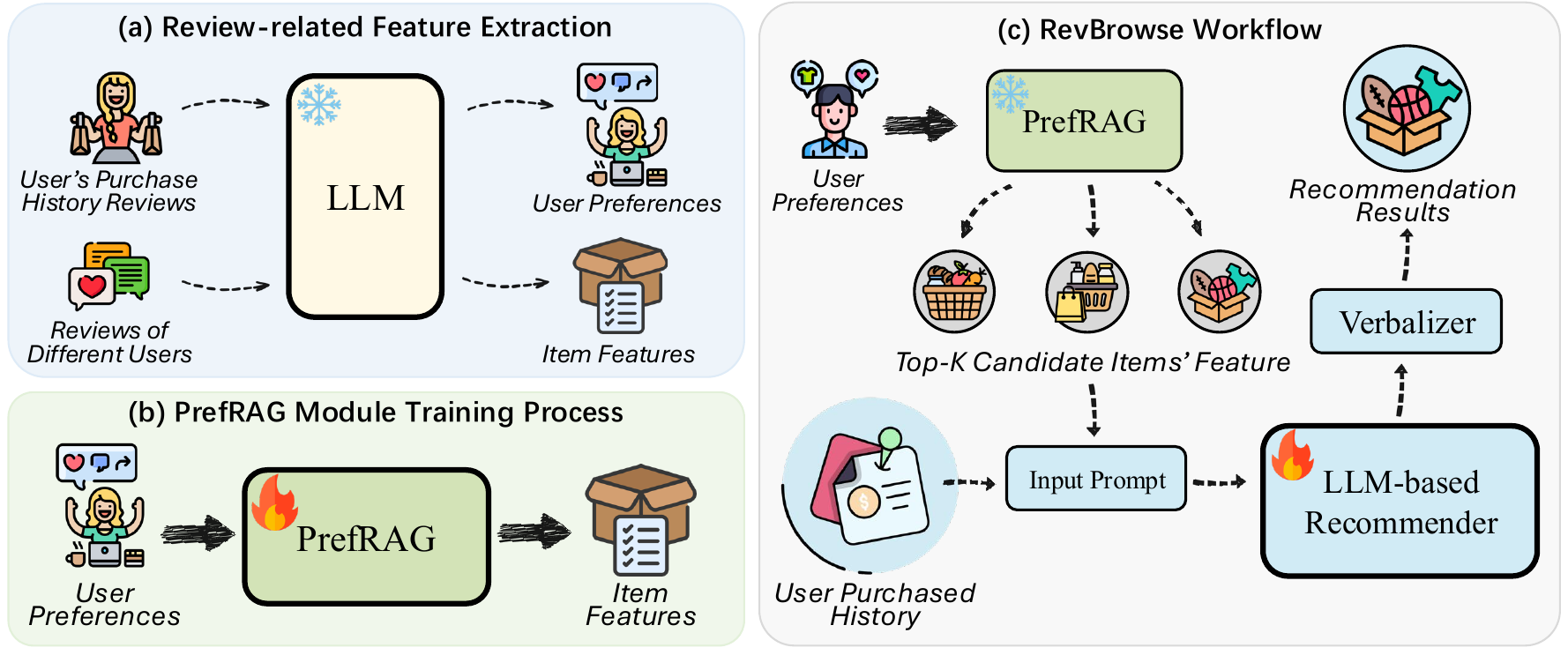}}
\caption{The details of the proposed PrefRAG module and RevBrowse framework: (a) An LLM extracts user preferences and item features from the user's review history and peer reviews; (b) The prefRAG module is trained with contrastive learning to retrieve item features with user preferences; (c) After training, prefRAG is frozen, and its outputs, together with purchase history, are fed into the LLM-based recommender for final prediction.}
\label{method}
\end{figure*}

\vspace{-10pt}
\subsection{Retrieval Augmented Generation}
In recent years, Retrieval-Augmented Generation (RAG) has emerged as a powerful paradigm to enhance Large Language Models (LLMs) by incorporating external knowledge into the generation process. By decoupling retrieval and generation, RAG enables models to access up-to-date and domain-specific information, thereby mitigating issues such as hallucination and outdated internal knowledge~\cite{DBLP:conf/icml/BorgeaudMHCRM0L22, DBLP:conf/nips/LewisPPPKGKLYR020}. Early frameworks like REALM~\cite{DBLP:journals/corr/abs-2002-08909} and FiD~\cite{DBLP:conf/eacl/IzacardG21} laid the foundation by integrating dense retrievers and encoder-decoder generators, achieving strong performance in knowledge-intensive tasks. Subsequent developments have focused on improving retrieval granularity~\cite{DBLP:conf/kdd/FanDNWLYCL24}, and improved adaptive augmentation strategies in models such as In-Context RALM~\cite{DBLP:journals/tacl/RamLDMSLS23} and SKR~\cite{DBLP:conf/emnlp/WangLSL23}. Additionally, novel training strategies, such as FiD-KD~\cite{DBLP:conf/iclr/IzacardG21} and REPLUG~\cite{DBLP:conf/naacl/ShiMYS0LZY24}, have enhanced model alignment and retrieval effectiveness.
Recently, there is a trend to integrate RAG into recommender systems. For example, CoRAL~\cite{DBLP:conf/kdd/WuC0HWHM24, DBLP:journals/jair/WeiDZWHL25} proposed to retrieve additional information (e.g., user profiles, user-item interactions) as collaborative information. Additionally, some work \cite{DBLP:journals/corr/abs-2501-02226, DBLP:conf/www/LiZLCRKL25} seek to apply RAG on knowledge graphs to migrate noise and incorporate structural relationships in knowledge. 

Although those RAG-based methods on RS have shown promise, they often rely on structured inputs like IDs or short descriptions, underutilizing rich user-generated content such as reviews. This limits their ability to model fine-grained, contextual user preferences. Moreover, user intent is typically treated as static, reducing adaptability across items. To address this, inspired by users’ “browsing-to-decision” behavior, we propose RevBrowse, a review-centric RAG-based recommendation framework that performs retrieval over user histories. By dynamically selecting relevant review-related features, RevBrowse enables personalized and explainable recommendation grounded in natural language.

\section{Preliminaries}
We consider a personalized recommendation scenario where each user $u \in U$ has an interaction–review history represented as $H_u = [(i_1, r_1), (i_2, r_2), \dots, (i_n, r_n)]$, where $i_k \in I$ is an item that the user has interacted with and $r_i$ is the corresponding review text. In addition, each candidate item $i \in I$ is associated with a set of reviews $R_i$ written by users who have interacted with that item. These textual reviews provide rich semantic information for modeling fine-grained user preferences and item characteristics.

Given the user’s historical sequence $H_u$ and the review corpus of candidate items ${\{R_i\}}_{i\in I}$, the goal is to recommend the next item for the user such that the recommended item aligns with the ground-truth next interaction $y_u^*$. To better capture dynamic preferences and enhance next-item prediction accuracy, we propose RevBrowse, a framework that utilizes a retrieval-augmented generation (RAG) to model reviews.

\section{Methodology}
In this section, we introduce the  proposed RevBrowse framework in details. As illustrated in Fig.~\ref{method}, there are three main components in RevBrowse  framework: (1) \textit{User Preference and Item Feature Extractor}, which derives user preferences and item features from review texts; (2) \textit{PrefRAG}, which implements a retrieval strategy to match item  features based on user preferences; and (3) a \textit{LLM-based Recommender}, which leverages a large language model to perform final recommendation over the candidate items.

\vspace{-10pt}
\subsection{Review-related Feature Extraction}
Recommender systems recommend different items to users based on their individual preferences, which are crucial for effective recommendations. In our method, as shown in Fig.~\ref{method}~(a),  we explicitly extract user preferences and item features using an LLM to represent user preferences. Below we describe how user preferences and item features are extracted.





\noindent\begin{minipage}{0.48\textwidth}
\vspace{10pt}
\begin{prompt}[title={User Preferences Extraction Prompt Template}] %
\textbf{Instruction:} \\
Given a list of items a user bought along with their title, reviews, and score in JSON format, generate user preferences.

\textbf{Input:} \\
User reviews: \{reviews\}

\textbf{Response:} \\
A JSON object with two keys: \texttt{Like} and \texttt{Dislike}... 
... 


\textbf{Output format:} \\
...
\end{prompt}
\captionof{table}{The user preferences extraction prompt template.}
\label{user_prompt_template}
\end{minipage}

\textbf{User Preferences Extraction.} The review data of a user’s historical purchases (denoted as $R_u = [r_1^u, r_2^u, \dots, r_n^u]$) reflects their preference tendencies. These tendencies could be categorized into two emotional dimensions: liked preferences (positive preferences) and disliked preferences (negative preferences). Liked preferences capture the item features that users appreciate as revealed through positive reviews, whereas disliked preferences encompass attributes users negatively perceive, indicated by negative feedback. We construct a prompt incorporating user reviews to enable the LLM to infer user preference tendencies, formally expressed as:
\begin{equation}
    t_{like}^u, t_{dislike}^u = \phi_t(R_u, T_{user}),
\end{equation}
here, $t_{like}^u$ and $t_{dislike}^u$ represent textual descriptions of the user's liked and disliked preferences, respectively. We use Qwen2.5-72B-instruct~\cite{qwen2.5}, denoted by $\phi_t$, for extracting user preferences. $R_u$ represents the user's historical reviews, and $T_{user}$ is a template for the user preference extraction task. The prompt template is shown in Table~\ref{user_prompt_template}. 
Please refer to Table~\ref{full_user_prompt_template} in  Appendix~\ref{appendix:Detailed Prompt Templates} for detailed prompt.






\noindent\begin{minipage}{0.48 \textwidth} 
\vspace{10pt}
\begin{prompt}[title={Item Features Extraction Prompt  Template}]
\textbf{Instruction:} \\
Given user reviews of purchased items in JSON format, extract high-level item properties from the comments.

\textbf{Input:} \\
Item review: \{review\}

\textbf{Response:} \\
A JSON object with two keys: \texttt{Pros} and \texttt{Cons}, ... 
... 


\textbf{Output format:} \\
...
\end{prompt}
\captionof{table}{The item features extraction prompt  template.}
\label{product Prompt Template}
\end{minipage}

\textbf{Item Features Extraction.} Different users provide varied reviews for the same item based on their preferences, allowing us to extract distinct item features from user reviews. We categorize item features into pros and cons, where pros are attributes favored by users and cons are features criticized by users. This extraction is formally expressed as:
\begin{equation}
    pros_k^i, cons_k^i = \phi_t(r_k^i, T_{item}),
\end{equation}
here, $pros_k^i$ and $cons_k^i$ represent textual descriptions of the pros and cons of item $i$. The variable $r_k^i$ denotes the $k$-th review of item $i$ ($i \in I$, $r_k^i \in R_i$), and $T_{item}$ is a template for the item feature extraction task. 
The prompt template is shown in Table~\ref{product Prompt Template}. 
Please refer to  Appendix~\ref{appendix:Detailed Prompt Templates} for the detailed prompt and example in Appendix E.

\begin{figure}[t]
\small
\centerline{\includegraphics[width=0.48\textwidth]{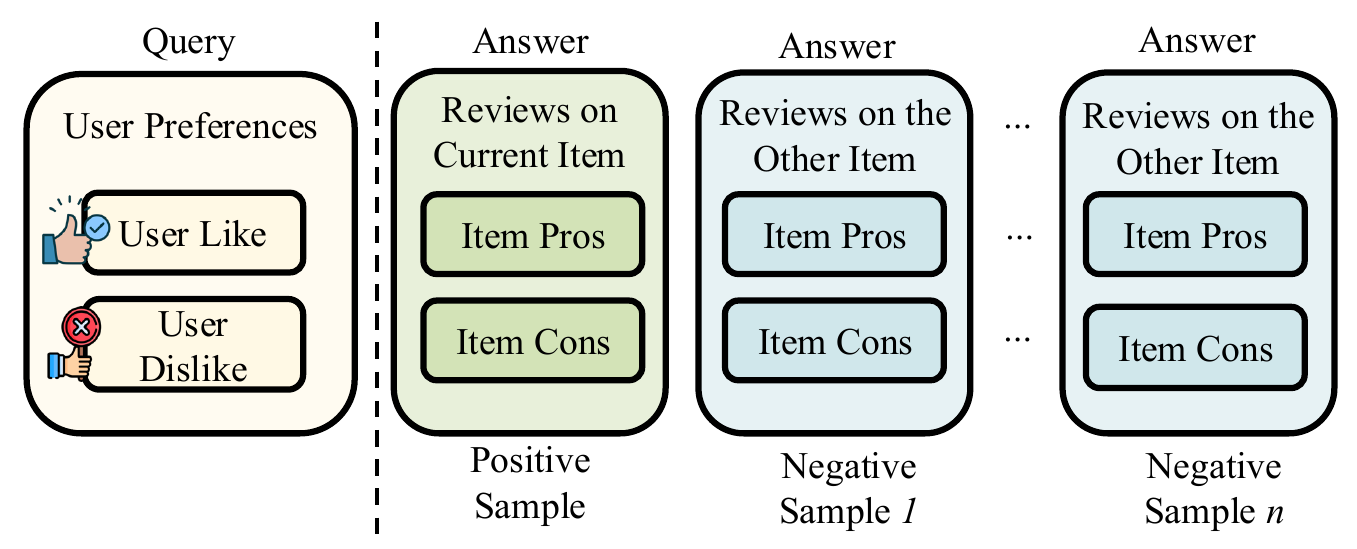}}
\caption{
Illustration of (a) the dual-tower architecture and (b) the workflow of contrastive learning in PrefRAG. Only the user-to-item retrieval branch for positive preferences (like-to-pros) is shown; the user-to-item retrieval branch for negative preferences (dislike-to-cons) is omitted for clarity.
}
\label{fig:CL}
\vspace{-10pt}
\end{figure}

\subsection{PrefRAG} 
When browsing an online shopping platform, users often focus on item reviews that match their preferences, particularly by examining the pros and cons of items before making a purchase decision.
Building upon the review-related feature extraction process, we draw an analogy between user preferences and item features: \textit{Like} and \textit{Dislike} represent user preferences, while \textit{Pros} and \textit{Cons} capture the positive and negative features of a item, respectively. 

To model this review-browsing behavior, we propose PrefRAG, a LLM-based feature retrieval model trained using contrastive learning (CL). As shown in Fig.~\ref{fig:CL}~(a), PrefRAG adopts a dual-tower architecture: user preferences and item features are encoded separately. In the experiment, the item features are pre-indexed to enable efficient retrieval. 

In addition, as illustrated in Fig.~\ref{method}~(b), PrefRAG’s primary objective is to learn the semantic relationships between user preferences and item features, thereby enabling the accurate retrieval of features that align with user preferences. Specifically, PrefRAG treats user preferences (i.e., \textit{Like/Dislike}) as queries and item features (i.e., \textit{Pros/Cons}) as answers.
When user preference is \textit{Like}, the model retrieves relevant item \textit{Pros}, and when input the \textit{Dislike} preference, it retrieves corresponding item \textit{Cons}. 
The model computes the semantic similarity between them and retrieves the top-K most relevant features. 

\textbf{Contrastive Learning Training Set Construction.}
The CL objective in PrefRAG aims to train the model to pull semantically aligned <\textit{user preference, item feature}> pairs closer (positive samples) while pushing apart irrelevant pairs (negative samples). 
Specifically, given a user's historical review sequence $R_u = [r_1^u, r_2^u, \dots, r_n^u]$ containing $n$ reviews, we adopt a sliding window approach to construct multiple contrastive training samples. The use of a sliding window in training sample construction allows PrefRAG to capture dynamically shifts in user preferences, ensuring that the model remains sensitive to evolving preferences over time rather than relying on static representations. For each window, we construct the positive samples and the negative samples for CL as follows:

First, to construct positive sample, the last review within the window, denoted as $r_n^u$, corresponds to a newly interacted item. From this item's review, we extract its pros and cons to serve as the positive answer $a^+$, reflecting the user's real evaluation of the item.

Next, to construct negative samples, we collect $m$ reviews written by \textit{other users} for the same item. Although these reviews describe the same item, they often reflect differing viewpoints. As a result, we serve their pros and cons as negative answers $a^-$ from the perspective of the target user.

\textbf{Contrastive Learning Loss.}
To explicitly align user preferences with item-level features, we decouple the two preference signals: 
we formulate \textit{likes} as queries for retrieving item \textit{pros}, and \textit{dislikes} as queries for retrieving item \textit{cons}. 
Below we take the branch of using user \textit{likes} to retrieve item \textit{pros} as an example. 
As shown in Fig.~\ref{fig:CL}~(b), 
given user preferences (\textit{Like}) as queries $q$ and item features (\textit{Pros}) as  answers $a$, 
we employ an LLM-based retriever $f_{\text{PrefRAG}}$ to obtain the last hidden-layer representations 
$h_q=[h_1^q, h_2^q, \dots, h_{l_q}^q]$ and $h_a=[h_1^a, h_2^a, \dots, h_{l_a}^a]$, 
where $l_q$ and $l_a$ denote the sequence lengths of $q$ and $a$, respectively. 
We then represent the query and answer by the embeddings of the final $[EOS]$ token, 
i.e., $e_q = h_{l_q}^q$ and $e_a = h_{l_a}^a$, where $e_q, e_a \in \mathbb{R}^d$. 
Given a batch of $B$ positive pairs $\{(q_i, a_i^+)\}_{i=1}^B$ and $m$ negative answer $\{a_{i,j}^-\}_{j=1}^m$ for each query, 
the cosine similarity between query and answer embeddings is computed as:
\begin{equation}
    s(q,a) = \frac{e_q^\top e_a}{\| e_q \| \, \| e_a \|}.
\end{equation}

Finally, we adopt the InfoNCE loss~\cite{DBLP:journals/corr/abs-1807-03748} to encourage queries (e.g., user likes) 
to be closer to their matched item features (e.g., positive samples' pros) while pushing away irrelevant ones (e.g., negative samples' pros):
\begin{equation}
    L = -\frac{1}{B} \sum_{i=1}^{B} 
    \log 
    \frac{\exp(s(q_i,a_i^+))}
    {\exp(s(q_i,a_i^+)) + \sum_{j=1}^{m} \exp(s(q_i,a_{i,j}^-))}.
\end{equation}

The same process is applied in the \textit{dislike} to \textit{cons} branch, 
where user dislikes serve as queries and item cons as answers.

\subsection{LLM-based Recommender}





The workflow is depicted in Fig.~\ref{method}~(c). 
Upon completion of the PrefRAG module's training, we freeze its parameters and subsequently employ it for item feature retrieval. Following~\cite{DBLP:journals/corr/abs-2311-02089}, we employ LRU~\cite{DBLP:journals/corr/abs-2310-02367} to get the initial candidate item set. We denote the sets of pros and cons extracted from the historical reviews of item $k$ as $pros_k$ and $cons_k$, respectively.  By utilizing the trained retriever $f_{PrefRAG}$, we generate matrix representations for their features:
\begin{equation}
    E_{pros}^k = f_{PrefRAG}(pros_k),
\end{equation}
\begin{equation}
    E_{cons}^k = f_{PrefRAG}(cons_k).
\end{equation}

To retrieve item features that align with a user's preferences, we first generate vector representations of their preferences:
\begin{equation}
    e_{like}^u = f_{PrefRAG}(t_{like}^u),
\end{equation}
\begin{equation}
    e_{dislike}^u = f_{PrefRAG}(t_{dislike}^u).
\end{equation}

We then compute the cosine similarity $cos(\cdot)$ 
between the user’s preference vector and each candidate item feature in the corresponding item feature set. 
Specifically, for a user preference vector (e.g., $e_{like}^u$) and a matrix of candidate embeddings (e.g., $E_{pros}^k$), 
$cos(\cdot)$ computes the cosine similarity between the vector and each row of the matrix individually. 
The top-$K$ most similar textual candidates are then retrieved:
\begin{equation}
    pros_k^* = \text{topK}\big( cos(e_{like}^u, E_{pros}^k) \big),
\end{equation}
\begin{equation}
    cons_k^* = \text{topK}\big( cos(e_{dislike}^u, E_{cons}^k) \big),
\end{equation}
where $\text{topK}(\cdot)$ returns the top-$K$ ranked texts based on similarity scores, $pros_k^*$ and $cons_k^*$ represent the retrieved text pros and cons that are most relevant to the user's expressed preferences.






After retrival, to enable preference-aware recommendation, we design a structured prompt that encapsulates the context required for personalized recommendation. 
As shown in Table~\ref{recommendation_prompt_template}, the input prompt contains  three key components: \texttt{history}, \texttt{preference}, and \texttt{candidates}, where \texttt{history} field contains the titles of items previously purchased by the user. The \texttt{preference} field extracts the user's preferences inferred from their historical reviews, which include two components: \texttt{Like} and \texttt{Dislike}. The \texttt{candidates} field contains the set of items retrieved for the next-item recommendation, for example:
(A) \textit{title}: Candy Cigarette, \textit{pros}: nostalgic taste, \textit{cons}: not a real smoking experience; (B) \textit{title}: Lenny \&amp;, \textit{pros}: nutritious, \textit{cons}: none noted ... Each candidate consists of a title, pros, and cons, where the pros and cons are retrieved from the sets $pros_k^*$ and $cons_k^*$ respectively. The \texttt{label} indicates the ground-truth index (in letter form) of the recommended item.

\noindent\begin{minipage}{0.48\textwidth}
\vspace{10pt}
\begin{prompt}[title={The Recommendation Input Prompt}]
\textbf{Instruction:} \\
Given user history in chronological order, recommend an item from the candidate pool with its index letter.

\textbf{Input:} \\
User history: \{history\}; \\
User preference: \{preference\}; \\
Candidate pool: \{candidates\}

\textbf{Response:} \\
\{label\}
\end{prompt}
\captionof{table}{The recommendation input prompt.}
\label{recommendation_prompt_template}
\end{minipage}

\begin{table}[t]
\centering
\small
\caption{Dataset statistics after preprocessing.}
\begin{tabular}{cccccc}
\toprule
\textbf{Datasets} & \textbf{Users} & \textbf{Items} & \textbf{Interact} & \textbf{Length} & \textbf{Density} \\
\midrule
Games   & 15,264  & 7,676   & 148K  & 9.69   & $1\mathrm{e}{-3}$ \\
Food & 14,681     & 8,713   & 151K  & 10.30 & $1\mathrm{e}{-3}$ \\
Clothing  & 39,387  & 23,033  & 278K  & 7.08   & $3\mathrm{e}{-4}$ \\
Sport   & 35,598  & 18,357   & 296K  & 8.32   & $4\mathrm{e}{-4}$ \\
\bottomrule
\end{tabular}
\label{tab:dataset_statistics}
\end{table}

With the prompt template constructed, and following the methodology proposed in LLaMARec~\cite{DBLP:journals/corr/abs-2311-02089}, we leverage LLaMA2-7B~\cite{DBLP:journals/corr/abs-2307-09288} as the LLM-based recommendation model. The model is trained to sort candidate items retrieved during the recall phase. We adopt the standard next-token prediction paradigm, where the \texttt{label} token is treated as the target item. Model parameters are optimized by maximizing the likelihood of correctly predicting the index of the ground-truth item: 
\begin{equation}
    \mathcal{L}_{\text{rec}} = - \frac{1}{N} \sum_{i=1}^{N} \log P_{\theta}(y_i \,|\, x_i),
\end{equation}
where $N$ is the number of training instances, 
$x_i$ denotes the input prompt, 
$y_i$ is one of the ground-truth item token, 
and $P_{\theta}(y_i \,|\, x_i)$ is the probability assigned by the model with parameters $\theta$. 


\begin{table*}[ht!]
\centering
\caption{Performance comparison of different methods on \textbf{Games} and \textbf{Food} datasets.}
\label{tab:perf_games_food}
\resizebox{0.9\textwidth}{!}{
\begin{tabular}{l|ccc|ccc|ccc|ccc}
\toprule
\multirow{2}{*}{\textbf{Method}} 
& \multicolumn{6}{c|}{\textbf{Games}} 
& \multicolumn{6}{c}{\textbf{Food}} \\
\cmidrule(lr){2-7} \cmidrule(lr){8-13}
& R@5 & N@5 & M@5 & R@10 & N@10 & M@10 & R@5 & N@5 & M@5 & R@10 & N@10 & M@10 \\
\midrule
LightGCN     & 0.0510 & 0.0345 & 0.0336 & 0.0844 & 0.0459 & 0.0390 & 0.0340 & 0.0234 & 0.0229 & 0.0557 & 0.0309 & 0.0267 \\
BERT4REC     & 0.0767 & 0.0468 & 0.0371 & 0.1304 & 0.0640 & 0.0440 & 0.0486 & 0.0308 & 0.0250 & 0.0791 & 0.0406 & 0.0290 \\
SASRec       & 0.0744 & 0.0467 & 0.0377 & 0.1226 & 0.0622 & 0.0440 & 0.0555 & 0.0371 & 0.0310 & 0.0825 & 0.0458 & 0.0346 \\
RGCL         & 0.0703     & 0.0402     & 0.0305     & 0.1284     & 0.0589     & 0.0381     & 0.0279     & 0.0157     & 0.0118     & 0.0615    & 0.0264     & 0.0161 \\
DeepCoNN     & 0.0692     & 0.0406     & 0.0313     & 0.1255     & 0.0586     & 0.0387  & 0.0250     & 0.0142     & 0.0107     & 0.0573     & 0.0245     & 0.0148   \\
NARRE        & 0.0632 & 0.0353 & 0.0263 & 0.1207 & 0.0537 & 0.0338 & 0.0262 & 0.0154 & 0.0119 & 0.0586 & 0.0256 & 0.0160 \\
LRU          & 0.0938 & 0.0620 & 0.0516 & 0.1398 & 0.0817 & 0.0602 & 0.0611 & 0.0418 & 0.0354 & 0.0887 & 0.0507 & 0.0391 \\
ZS-Ranker    & 0.0942 & 0.0621 & 0.0516 & 0.1456 & 0.0785 & 0.0583 & 0.0608 & 0.0386 & 0.0314 & 0.0903 & 0.0480 & 0.0352 \\
LlamaRec     & 0.0999 & 0.0647 & 0.0532 & 0.1527 & 0.0817 & 0.0602 & 0.0636 & 0.0425 & 0.0356 & 0.0945 & 0.0524 & 0.0396 \\
Exp3rt       & 0.1188 & 0.0809 & 0.0685 & 0.1670 & 0.0965 & 0.0749 & 0.0639 & 0.0443 & 0.0379 & 0.0971 & 0.0550 & 0.0422 \\
RevBrowse (ours) 
             & \textbf{0.1219} & \textbf{0.0845} & \textbf{0.0722} 
             & \textbf{0.1680} & \textbf{0.0994} & \textbf{0.0783} 
             & \textbf{0.0657} & \textbf{0.0464} & \textbf{0.0401} 
             & \textbf{0.0987} & \textbf{0.0571} & \textbf{0.0445} \\
\bottomrule
\end{tabular}
}
\end{table*}

\begin{table*}[ht!]
\centering
\small
\caption{Performance comparison of different methods on \textbf{Sport} and \textbf{Clothing} datasets.}
\label{tab:perf_sport_clothing}
\resizebox{0.9\textwidth}{!}{
\begin{tabular}{l|ccc|ccc|ccc|ccc}
\toprule
\multirow{2}{*}{\textbf{Method}} 
& \multicolumn{6}{c|}{\textbf{Sport}} 
& \multicolumn{6}{c}{\textbf{Clothing}} \\
\cmidrule(lr){2-7} \cmidrule(lr){8-13}
& R@5 & N@5 & M@5 & R@10 & N@10 & M@10 & R@5 & N@5 & M@5 & R@10 & N@10 & M@10 \\
\midrule
LightGCN     & 0.0285 & 0.0187 & 0.0177 & 0.0445 & 0.0241 & 0.0204 & 0.0175 & 0.0112 & 0.0096 & 0.0285 & 0.0148 & 0.0112 \\
BERT4REC     & 0.0217 & 0.0134 & 0.0107 & 0.0407 & 0.0195 & 0.0132 & 0.0110 & 0.0072 & 0.0060 & 0.0191 & 0.0098 & 0.0070 \\
SASRec       & 0.0255 & 0.0162 & 0.0131 & 0.0427 & 0.0217 & 0.0154 & 0.0102 & 0.0066 & 0.0054 & 0.0186 & 0.0093 & 0.0065 \\
RGCL         & 0.0176     & 0.0097     & 0.0072     & 0.0392     & 0.0166     & 0.0100     & 0.0091     & 0.0050     & 0.0037     & 0.0207     & 0.0087     & 0.0052  \\
DeepCoNN     & 0.0164     & 0.0093     & 0.0070     & 0.0360     & 0.0155     & 0.0095     & 0.0087     & 0.0050     & 0.0038     & 0.0201     & 0.0086     & 0.0052  \\
NARRE        & 0.0179 & 0.0103 & 0.0078 & 0.0372 & 0.0164 & 0.0103 & 0.0085 & 0.0047 & 0.0034 & 0.0199 & 0.0083 & 0.0049 \\
LRU          & 0.0341 & 0.0244 & 0.0212 & 0.0496 & 0.0294 & 0.0233 & 0.0185 & 0.0136 & 0.0106 & 0.0271 & 0.0153 & 0.0117 \\
ZS-Ranker    & 0.0336 & 0.0209 & 0.0167 & 0.0501 & 0.0261 & 0.0188 & 0.0186 & 0.0120 & 0.0099 & 0.0287 & 0.0153 & 0.0112 \\
LlamaRec     & 0.0359 & 0.0247 & 0.0211 & 0.0528 & 0.0302 & 0.0233 & 0.0206 & 0.0141 & 0.0119 & 0.0289 & 0.0167 & 0.0130 \\
Exp3rt       & 0.0353 & 0.0249 & 0.0215 & 0.0528 & 0.0305 & 0.0238 & 0.0216 & 0.0144 & 0.0120 & 0.0304 & 0.0173 & 0.0132 \\
RevBrowse (ours) 
             & \textbf{0.0369} & \textbf{0.0264} & \textbf{0.0230} 
             & \textbf{0.0544} & \textbf{0.0320} & \textbf{0.0253} 
             & \textbf{0.0227} & \textbf{0.0164} & \textbf{0.0143} 
             & \textbf{0.0308} & \textbf{0.0190} & \textbf{0.0154} \\
\bottomrule
\end{tabular}
}
\end{table*}

Finally, we utilize the trained LLM-based recommender for inference. The prompt is presented without the ground-truth label, and the model outputs a probability distribution over the candidate indices. To ensure the model generates valid  outputs, we incorporate a verbalizer, which defines a mapping between candidate items and their corresponding index tokens (e.g., A, B, C, ...). This verbalizer constrains the model’s output space to a closed set of predefined index letters, enabling consistent and reliable predictions.
The logits associated with each letter index (i.e., \texttt{logits}(A), \texttt{logits}(B), ..., \texttt{logits}(T)) are interpreted as recommendation scores, and are used to rank the candidate items accordingly. In this way, the sequential recommendation process is effectively formulated as a classification task over a fixed set of item indices, guided by the structure imposed by the verbalizer.




\section{Experiment}
\subsection{Experiment Settings}
\subsubsection{Dataset}
We conduct sequential recommendation experiments on four subsets of the Amazon-2014 dataset\footnote{https://jmcauley.ucsd.edu/data/amazon/.}: \textit{Food}, \textit{Sports}, \textit{Clothing}, and \textit{Games}, to evaluate the effectiveness of the proposed RevBrowse method. The Amazon-2014 dataset is a publicly available large-scale item review dataset released by Amazon, containing reviews and metadata across various item categories. In the experiments, we utilize both user review texts and item title information from the metadata.
To balance computational efficiency and data quality, following the preprocessing procedure in~\cite{DBLP:journals/corr/abs-2311-02089}, we filter out users and items with fewer than five interactions.
Detailed statistics of the datasets in the experiment are presented in Table~\ref{tab:dataset_statistics}.

\vspace{-10pt}
\subsubsection{Baselines}
To evaluate the effectiveness of the proposed RevBrowse, we compare it against a range of strong baselines from three categories: traditional Collaborative Filtering (CF) based methods without side information (\textit{i.e.}, LightGCN~\cite{DBLP:conf/sigir/0001DWLZ020}, BERT4REC~\cite{DBLP:conf/cikm/SunLWPLOJ19}, SASRec~\cite{DBLP:conf/icdm/KangM18}, LRU~\cite{DBLP:journals/corr/abs-2310-02367}), review-based recommendation methods (\textit{i.e.}, RGCL~\cite{DBLP:conf/sigir/ShuaiZWSHWL22}, DeepCoNN~\cite{DBLP:conf/wsdm/ZhengNY17}, NARRE~\cite{DBLP:conf/www/ChenZLM18}), and LLM-enhanced works (\textit{i.e.}, ZS-Ranker~\cite{DBLP:conf/sigir/Kim0CKCY025}, LlamaRec~\cite{DBLP:journals/corr/abs-2311-02089}, Exp3rt~\cite{DBLP:journals/corr/abs-2408-06276}). Detailed baseline descriptions are shown in the Appendix~\ref{app:baseline}.

\begin{table*}[htbp]
\centering
\small
\caption{Ablation study of different feature components across datasets.}
\label{tab:ablation_restructured}
\resizebox{0.9\textwidth}{!}{
\begin{tabular}{l|ccc|ccc|ccc|ccc}
\toprule
\multirow{2}{*}{\textbf{Method}} 
& \multicolumn{6}{c|}{\textbf{Games}} 
& \multicolumn{6}{c}{\textbf{Food}} \\
\cmidrule(lr){2-7} \cmidrule(lr){8-13}
& R@5 & N@5 & M@5 & R@10 & N@10 & M@10 & R@5 & N@5 & M@5 & R@10 & N@10 & M@10 \\
\midrule
\emph{w/o} User Pref. \& Reviews         & 0.0999 & 0.0647 & 0.0532 & 0.1527 & 0.0817 & 0.0602 & 0.0636 & 0.0425 & 0.0356 & 0.0945 & 0.0524 & 0.0396 \\
\emph{w/o} User Pref.     & 0.1175 & 0.0767 & 0.0633 & 0.1661 & 0.0924 & 0.0698 & 0.0651 & 0.0452 & 0.0387 & 0.0969 & 0.0554 & 0.0428 \\
\emph{w/o} Reviews    & 0.1014 & 0.0659 & 0.0542 & 0.1528 & 0.0823 & 0.0609 & 0.0642 & 0.0442 & 0.0376 & 0.0926 & 0.0533 & 0.0413 \\
RevBrowse (ours)   & \textbf{0.1219} & \textbf{0.0845} & \textbf{0.0722} & \textbf{0.1680} & \textbf{0.0994} & \textbf{0.0783} & \textbf{0.0657} & \textbf{0.0464} & \textbf{0.0401} & \textbf{0.0987} & \textbf{0.0571} & \textbf{0.0445} \\
\midrule
\multirow{2}{*}{} 
& \multicolumn{6}{c|}{\textbf{Clothing}} 
& \multicolumn{6}{c}{\textbf{Sport}} \\
\cmidrule(lr){2-7} \cmidrule(lr){8-13}
& R@5 & N@5 & M@5 & R@10 & N@10 & M@10 & R@5 & N@5 & M@5 & R@10 & N@10 & M@10 \\
\midrule
\emph{w/o} User Pref. \& Reviews       & 0.0206 & 0.0141 & 0.0119 & 0.0289 & 0.0167 & 0.0130 & 0.0359 & 0.0247 & 0.0211 & 0.0528 & 0.0302 & 0.0233 \\
\emph{w/o} User Pref.     & 0.0220 & 0.0160 & 0.0140 & 0.0303 & 0.0187 & 0.0151 & 0.0353 & 0.0249 & 0.0215 & 0.0528 & 0.0305 & 0.0238 \\
\emph{w/o} Reviews    & 0.0207 & 0.0149 & 0.0130 & 0.0298 & 0.0178 & 0.0142 & 0.0365 & 0.0244 & 0.0204 & 0.0543 & 0.0301 & 0.0227 \\
RevBrowse (ours)   & \textbf{0.0227} & \textbf{0.0164} & \textbf{0.0143} & \textbf{0.0308} & \textbf{0.0190} & \textbf{0.0154} & \textbf{0.0369} & \textbf{0.0264} & \textbf{0.0230} & \textbf{0.0544} & \textbf{0.0320} & \textbf{0.0253} \\
\bottomrule
\end{tabular}
}
\end{table*}

\subsubsection{Evaluation Metrics}
We adopt the leave-one-out evaluation protocol, where for each user sequence, the last interacted item is held out for testing, the second-to-last item is used for validation, and the remaining interactions are used for training. We evaluate the model performance using three standard metrics: Recall (R@$k$), Normalized Discounted Cumulative Gain (N@$k$), and Mean Reciprocal Rank (M@$k$), with $k \in [5, 10]$. During evaluation, predictions are ranked over the entire item set. For model selection, we retain the checkpoint achieving the best validation performance. 


%

\vspace{-10pt}
\subsubsection{Implementation Details}
We use frozen Qwen2.5-72B-instruct \cite{qwen2.5} for user preferences and item  features extraction. 
Following LlamaRec~\cite{DBLP:journals/corr/abs-2311-02089}, 
we employ LRU~\cite{DBLP:journals/corr/abs-2310-02367} to get the initial candidate item set, with default hyperparameters configuration in the original paper. 
Both PrefRAG and the LLM-based recommender use Llama-2-7b-hf~\cite{DBLP:journals/corr/abs-2307-09288} as the backbone and are trained with LoRA~\cite{DBLP:conf/iclr/HuSWALWWC22}, utilizing early stopping based on validation performance. In the PrefRAG model, reviews are processed with a sliding window size of 20 and a 1:40 positive-to-negative sample ratio per batch. The model is trained for a maximum of 5 epochs. During data processing, we remove reviews containing item features that are exactly identical to those of the positive samples. 
For the LLM-based recommender, the batch size is set to 1, with a history length of 20 items. The learning rate is set to 1e-4, and the optimal hyperparameters are selected through grid search.


\vspace{-9pt}
\subsection{Performance Comparison}
The experimental results on the four Amazon datasets (Table~\ref{tab:perf_games_food} and Table~\ref{tab:perf_sport_clothing}) show that RevBrowse consistently outperforms all baselines across all datasets and metrics, and we could find: 
(1) Compared with traditional CF-based methods, RevBrowse achieves clear and consistent improvements by leveraging review information in addition to user–item interactions. This demonstrates the importance of reviews, as CF signals alone are often too sparse to fully capture user preferences, while reviews provide richer semantic cues.
(2) Compared with review-based recommendation methods, RevBrowse can more accurately identify and utilize reviews that are most relevant to the target item. The PrefRAG module enables fine-grained review selection conditioned on the item, allowing the model to extract more dynamic user intent information.
(3) Against LLM-enhanced methods, RevBrowse demonstrates stronger capability to filter and organize large volumes of reviews. PrefRAG helps the LLM focus on the most relevant content, overcoming the context-length limitation and leading to more accurate and context-aware recommendations.

\begin{figure}[t!]
\centerline{\includegraphics[width=0.42\textwidth]{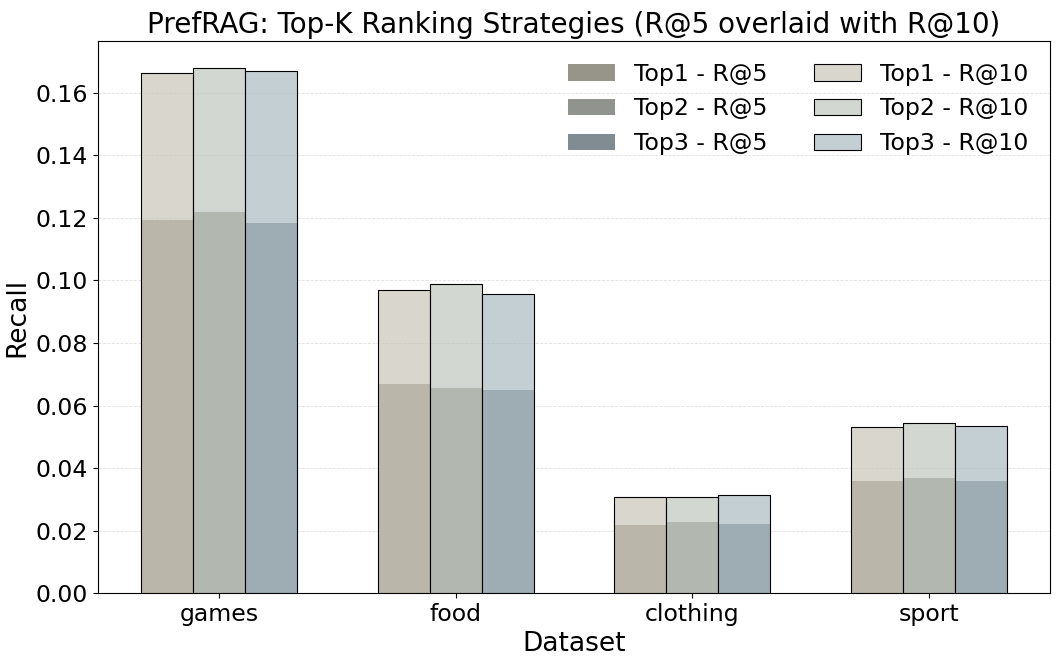}}
\caption{Performance of different top-K strategies in PrefRAG across different datasets. }
\label{fig:topk}
\end{figure}

\vspace{-10pt}
\subsection{Ablation Study}
To evaluate the contribution of different components in the input prompt (as illustrated in Table~\ref{recommendation_prompt_template}), we design three RevBrowse variants by selectively removing specific elements: \emph{w/o} User Pref. \& Reviews (removing both the user preference part and the PrefRAG-selected candidate reviews from the prompt), \emph{w/o} User Pref. (removing only the user preference part), and \emph{w/o} Reviews (removing only the PrefRAG-selected reviews).
As shown in Table~\ref{tab:ablation_restructured}, we could find that: 
(1) Removing only user preferences causes a moderate drop, suggesting that the global summary of user preferences provides complementary context but is less item-specific. 
(2) Removing only PrefRAG-selected reviews leads to a larger decline, especially on Games and Clothing datasets, highlighting that fine-grained review filtering supplies critical item-level evidence and helps the LLM focus on the most relevant information. 
(3) Removing both components yields the largest degradation, demonstrating that the two elements are highly complementary: user preferences offer a broad view of preference tendencies, while PrefRAG-selected reviews inject precise, context-specific signals. Their combination enables RevBrowse to balance global and fine-grained cues, achieving superior performance.

\begin{figure}[t]
\small
\centerline{\includegraphics[width=0.48\textwidth]{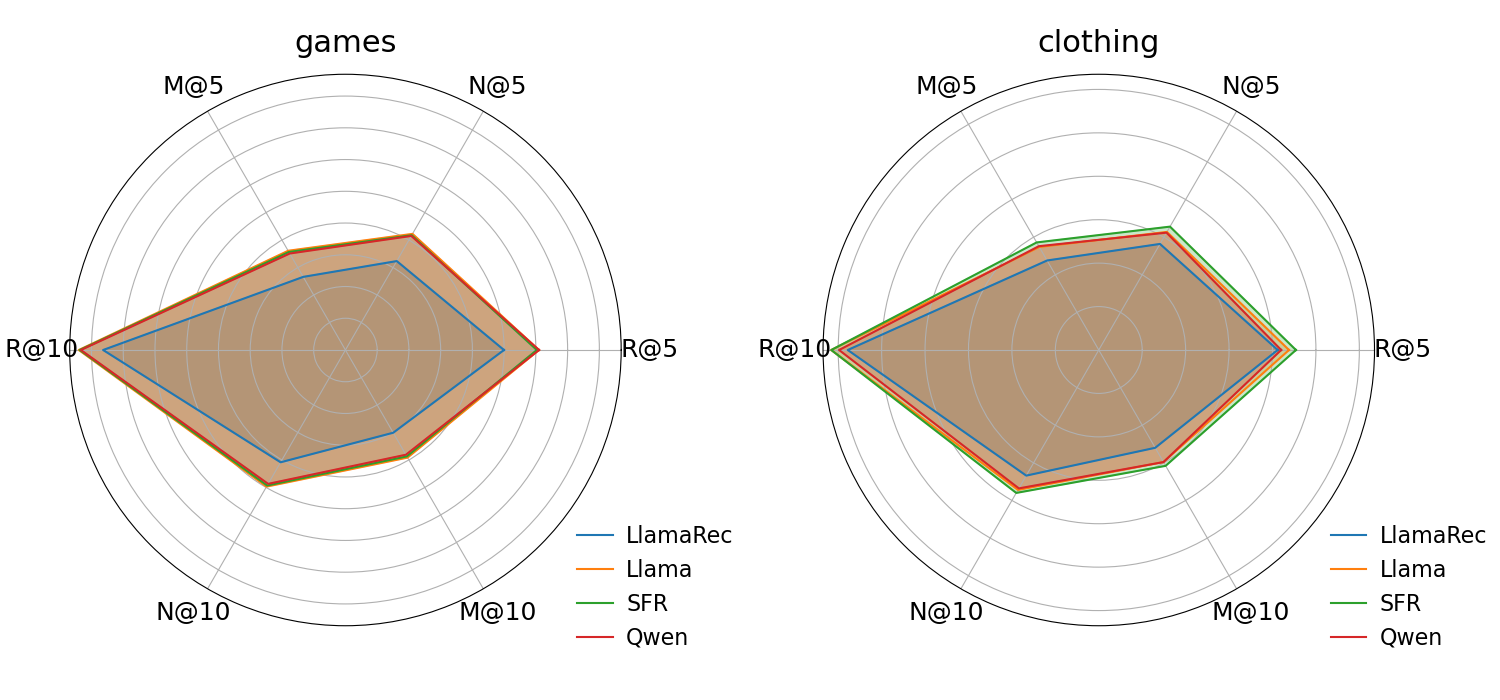}}
\caption{Performance comparison of different backbones of PrefRAG across Amazon Games and Clothing datasets. }
\label{fig:backbone}
\end{figure}

\subsection{Influence of Number of Reviews}
We investigate how the number of selected reviews in PrefRAG impacts recommendation performance by comparing top-1, top-2, and top-3 review strategies. Results in Fig.\ref{fig:topk} and Table\ref{appendix:top_k_comparison} (Appendix B) show that (1) Across all datasets, selecting the top-2 reviews consistently yields the best or near-best performance in terms of Recall, NDCG, and MRR, balancing relevant information and noise reduction.
(2) Selecting only the top-1 review generally underperforms, likely due to limited coverage of user preferences, as it does not capture enough diversity in the user's preferences.
(3) The top-3 review selection does not always improve performance and can sometimes introduce less relevant content, leading to a slight decrease in performance, particularly in the Games and Clothing datasets.
(4) In the Food dataset, although top-1 outperforms top-2 on Recall@5, top-2 performs better across most metrics, especially Recall@10 and MRR. Similarly, top-2 consistently provides the most balanced and stable improvements in the Sport dataset.
These results confirm that a small, high-quality set of relevant reviews is more beneficial than increasing the number of reviews, with PrefRAG’s selective review strategy enhancing recommendation accuracy.

\subsection{Influence of Different Backbones}
We evaluate the stability of PrefRAG under different LLM backbones by employing three alternatives: SFR~\cite{SFRAIResearch2024}, Qwen~\cite{qwen2.5}, and LLaMA~\cite{touvron2023llama}.
Fig.~\ref{fig:backbone} presents the results across Amazon games and clothing datasets,  Table~\ref{appedix:backbone_comparison} in Appendix B for more detail. We could find: 
(1) RevBrowse consistently outperforms LlamaRec with all three alternative backbones, confirming the model-agnostic nature of our framework.
(2) SFR and Qwen generally achieve better or comparable results to LLaMA, especially on the Food, Clothing, and Sport datasets. SFR performs best on Clothing and Sport, while Qwen excels on Food and Sport.
(3) LLaMA performs best on the Games dataset, suggesting that different backbones may capture user intent more effectively in different domains.
While the choice of backbone can impact the fine-grained quality of review understanding, RevBrowse maintains stable effectiveness across models, making it adaptable to various LLM backbones.





\begin{figure}[t]
\small
\centerline{\includegraphics[width=0.48\textwidth]{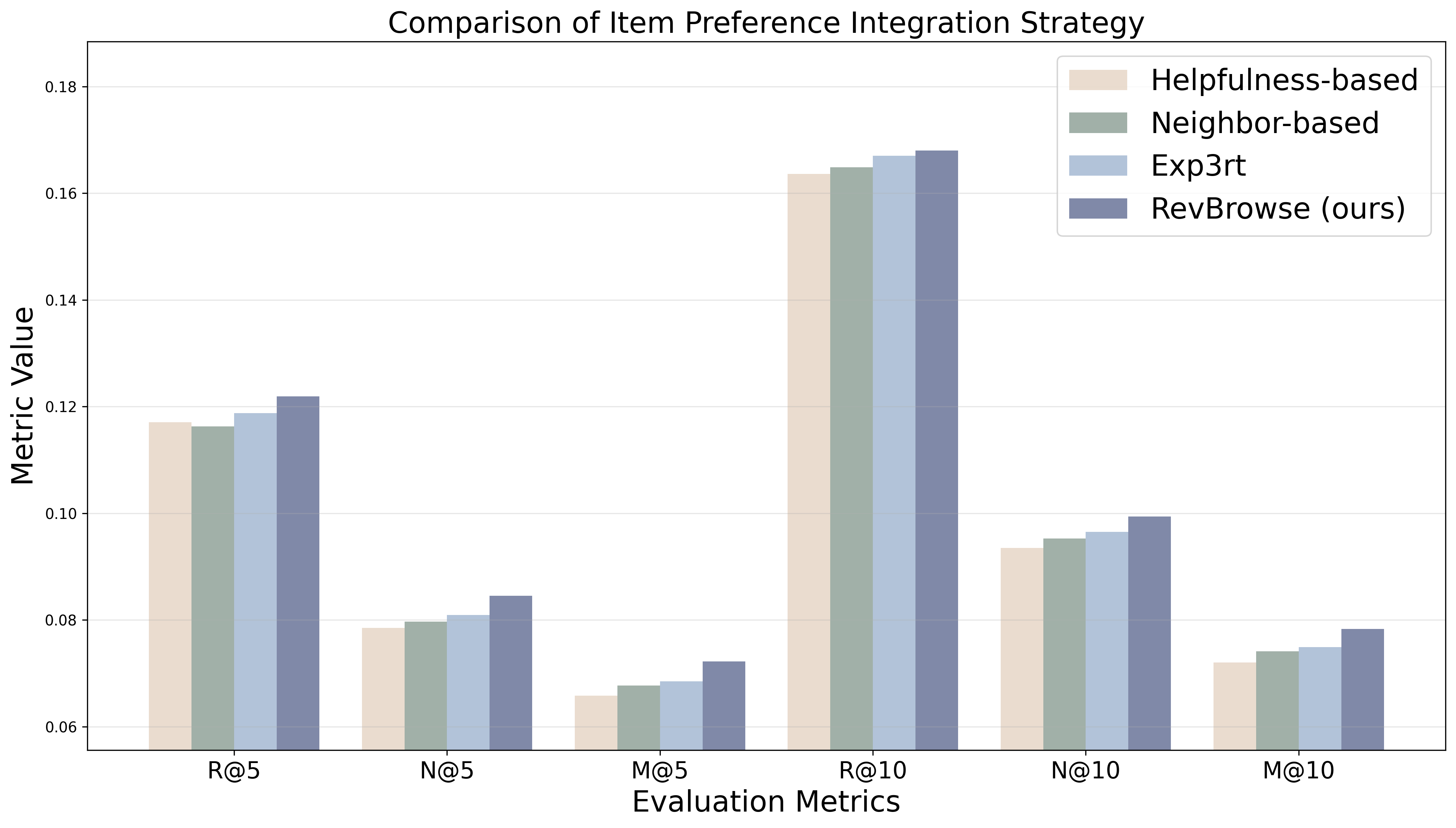}}
\caption{Performance of different user preference integration strategies in PrefRAG on Aamzon Games dataset. }
\label{fig:preference_integration}
\end{figure}

\subsection{Influence of Feature Integration Strategy}
In Fig.~\ref{fig:preference_integration}, we compare the RAG mechanism in RevBrowse with three alternative item feature integration strategies: Helpfulness-based (i.e., selecting only the
feature extracted from the reviews with the highest number of helpful votes), Neighbor-based (i.e., using the feature from the reviews written by the users with similar feature to a target user), and Exp3rt~\cite{DBLP:conf/sigir/Kim0CKCY025}. We could find: 
(1) RevBrowse achieves the best feature across all evaluation metrics, indicating the effectiveness of our feature integration strategy.
(2) While Helpfulness-based and Neighbor-based strategies perform similarly, they lag behind Exp3rt and RevBrowse due to limited ability to capture personalized and context-aware signals. 
(3) The largest improvements are observed on M@5 and M@10, where RevBrowse surpasses the best baseline (Exp3rt) by a clear margin. This highlights that selecting preference-relevant item features in a structured manner helps the model make more accurate top-ranked recommendations.

\vspace{-5pt}
\subsection{Case Study}
\begin{figure}[t]
\small
\centerline{\includegraphics[width=0.48\textwidth]{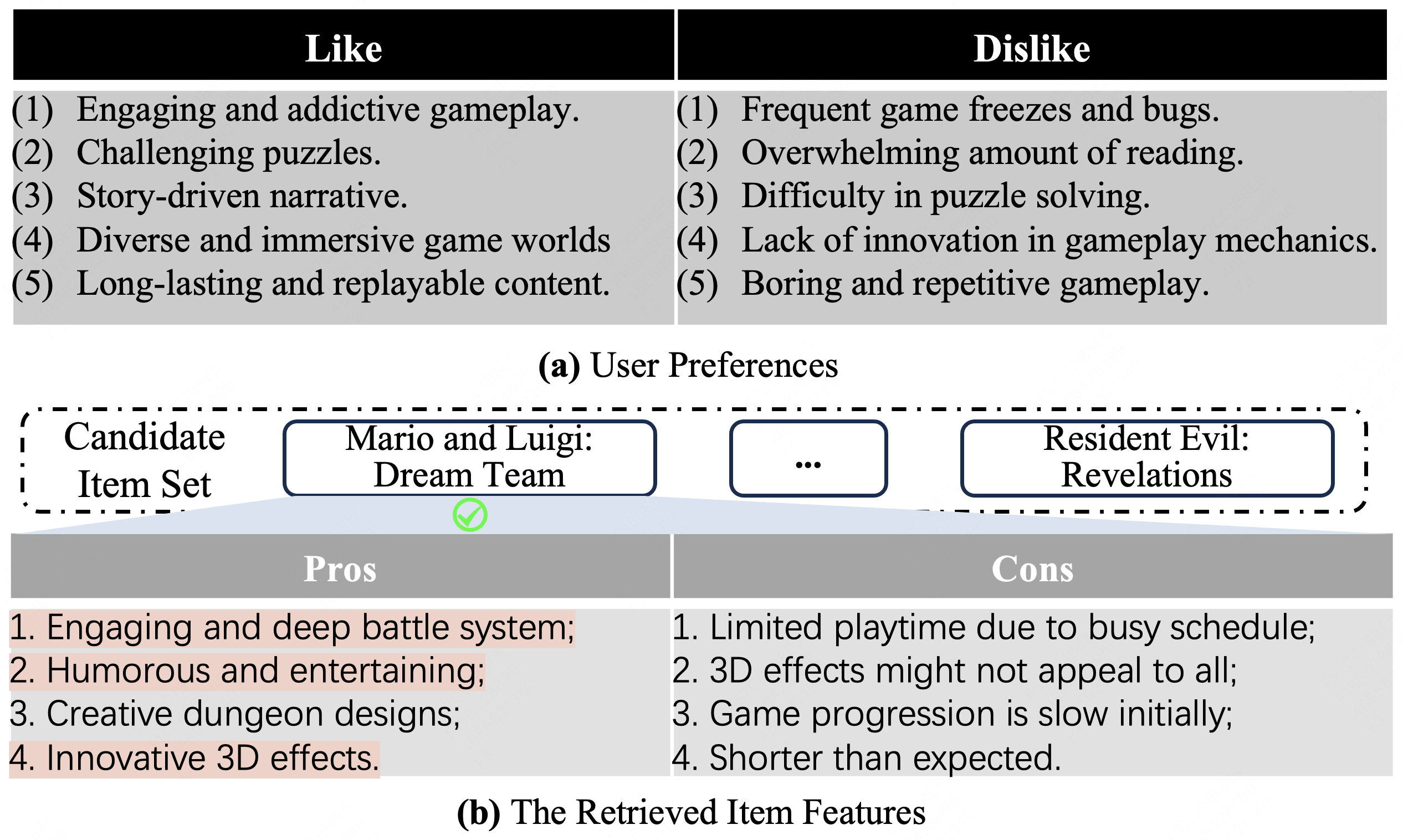}}
\caption{Case study of RevBrowse on the Amazon Games dataset. Given the user preferences in (a), the model first retrieves relevant item features from the candidate set as shown in (b), and subsequently makes prediction. }
\label{fig:case_study}
\end{figure}

To better illustrate the recommendation process of RevBrowse, we present a case study in Fig.~\ref{fig:case_study}. User likes and dislikes are first extracted to represent preference signals, while pros and cons for each candidate item are summarized   from user reviews. Due to input length constraints and the potential noise introduced by irrelevant content, we use PrefRAG to retrieve the most relevant item features based on user preferences.
For example, Fig.~\ref{fig:case_study}(b) shows 4 pros and 4 cons of item ‘\textit{Mario \& Luigi: Dream Team}’ retrieved by user preferences. Among them, 3 pros match the user's preferences, and none of the cons conflict with their dislikes, effectively capturing the user's preferences and enabling the model to correctly recommend the item. 
A full version of this case and the detailed data workflow of RevBrowse is shown in Fig.~\ref{appendix:case study all} in Appendix D.

\vspace{-5pt}
\section{Conclusion}
In this paper, inspired from the human “browsing-then-decision” behavior in online shopping, we propose RevBrowse, a review-driven  recommendation framework. 
By combining retrieval-augmented generation with sequential modeling, RevBrowse allows LLMs to dynamically utilize long, diverse user reviews while mitigating input length and noise issues. To enhance relevance and model focus, we introduce PrefRAG, a preference-aware RAG module that selects item features tailored to the target item. Experiments on four Amazon benchmarks show that RevBrowse not only outperforms strong LLM-based baselines, but also enhances recommendation interpretability through an explicit RAG process that makes the reasoning path transparent.
\bibliographystyle{ACM-Reference-Format}
\bibliography{sample-base}

\appendix

\begin{table*}[htbp]
\centering
\caption{Performance of top-K  strategies in PrefRAG across different datasets.}
\begin{tabular}{cccccccc}
\toprule
\textbf{Dataset} & \textbf{Methods} & \textbf{R@5} & \textbf{N@5} & \textbf{M@5} & \textbf{R@10} & \textbf{N@10} & \textbf{M@10} \\
\midrule

\multirow{3}{*}{games}
  & Top1 & 0.1192 & 0.0811 & 0.0686 & 0.1662 & 0.0963 & 0.0748 \\
  & Top2 & \textbf{0.1219} & \textbf{0.0845} & \textbf{0.0722} & \textbf{0.1680} & \textbf{0.0994} & \textbf{0.0783} \\
  & Top3 & 0.1182 & 0.0808 & 0.0685 & 0.1669 & 0.0966 & 0.0750 \\
\midrule

\multirow{3}{*}{food}
  & Top1 & \textbf{0.0668} & \textbf{0.0465} & 0.0398 & 0.0968 & 0.0562 & 0.0438 \\
  & Top2 & 0.0657 & 0.0464 & \textbf{0.0401} & \textbf{0.0987} & \textbf{0.0571} & \textbf{0.0445} \\
  & Top3 & 0.0651 & 0.0431 & 0.0359 & 0.0958 & 0.0530 & 0.0399 \\
\midrule

\multirow{3}{*}{clothing}
  & Top1 & 0.0219 & 0.0157 & 0.0137 & 0.0308 & 0.0186 & 0.0149 \\
  & Top2 & \textbf{0.0227} & \textbf{0.0164} & \textbf{0.0143} & 0.0308 & \textbf{0.0190} & \textbf{0.0154} \\
  & Top3 & 0.0221 & 0.0146 & 0.0121 & \textbf{0.0315} & 0.0176 & 0.0133 \\
\midrule

\multirow{3}{*}{sport}
  & Top1 & 0.0360 & 0.0257 & 0.0223 & 0.0530 & 0.0311 & 0.0245 \\
  & Top2 & \textbf{0.0369} & \textbf{0.0264} & \textbf{0.0230} & \textbf{0.0544} & \textbf{0.0320} & \textbf{0.0253} \\
  & Top3 & 0.0358 & 0.0235 & 0.0195 & 0.0535 & 0.0293 & 0.0218 \\
\bottomrule
\end{tabular}
\label{appendix:top_k_comparison}
\end{table*}

\section{Detailed Prompt Templates}
\label{appendix:Detailed Prompt Templates}

The detailed prompt template for user preferences extraction and item features extraction are shown in Table~\ref{full_user_prompt_template} and Table~\ref{full product Prompt Template}.

\label{prompt_templates}
\noindent\begin{minipage}{0.48\textwidth}
\vspace{10pt}
\begin{prompt}[title={User Preferences Extraction Prompt Template}]
\textbf{Instruction:} \\
Given a list of items a user bought along with their title, reviews, and score in JSON format, generate user preferences.

\textbf{Input:} \\
User reviews: \{reviews\}

\textbf{Response:} \\
A JSON object with two keys: \texttt{Like} and \texttt{Dislike}, each containing up to 5 high-level user preferences based on the comments. \\
Preferences must: \\
1. Reflect general likes and dislikes, not specific brands or items. \\
2. Be derived from the content of the reviews. \\
3. Exclude mentions of delivery time or pricing. \\
4. Be concise and simple. \\

\textbf{Output format:} \\
\begin{verbatim}
{
  "Like": ["..."],
  "Dislike": ["..."]
}
\end{verbatim}
\end{prompt}
\captionof{table}{The Detailed User Preference Extraction Prompt.}
\label{full_user_prompt_template}
\end{minipage}

\noindent\begin{minipage}{0.48 \textwidth} 
\label{itme_features_extraction}
\vspace{10pt}
\begin{prompt}[title={Item Features Extraction Prompt  Template}]
\textbf{Instruction:} \\
Given user reviews of purchased items in JSON format, extract high-level item properties from the comments.

\textbf{Input:} \\
Item review: \{review\}

\textbf{Response:} \\
A JSON object with two keys: \texttt{Pros} and \texttt{Cons}, each containing up to 5 high-level item properties. \\
Properties must: \\
1. Summarize general strengths and weaknesses of the items. \\
2. Be derived from the content of the reviews. \\
3. Avoid mentioning specific brands or item names. \\
4. Exclude any comments related to delivery time or pricing. \\
5. Be simple, short, and concise.

\textbf{Output format:} \\
\begin{verbatim}
{
  "Pros": ["..."],
  "Cons": ["..."]
}
\end{verbatim}
\end{prompt}
\captionof{table}{The Detailed Item Features Extraction Prompt.}
\label{full product Prompt Template}
\end{minipage}

\begin{table*}[htbp]
\centering
\caption{Performance comparison of different backbones of PrefRAG across different datasets.}
\begin{tabular}{cccccccc}
\toprule
\textbf{Dataset} & \textbf{Methods} & \textbf{R@5} & \textbf{N@5} & \textbf{M@5} & \textbf{R@10} & \textbf{N@10} & \textbf{M@10} \\
\midrule

\multirow{4}{*}{games}
  & w/o item pref & 0.1014 & 0.0659 & 0.0542 & 0.1528 & 0.0823 & 0.0609 \\
  & Llama    & 0.1219 & \textbf{0.0845} & \textbf{0.0722} & \textbf{0.1680} & \textbf{0.0994} & \textbf{0.0783} \\
  & SFR      & 0.1206 & 0.0835 & 0.0713 & 0.1674 & 0.0986 & 0.0774 \\
  & Qwen     & \textbf{0.1220} & 0.0830 & 0.0702 & 0.1667 & 0.0974 & 0.0762 \\
\midrule

\multirow{4}{*}{food}
  & w/o item pref & 0.0642 & 0.0442 & 0.0376 & 0.0926 & 0.0533 & 0.0413 \\
  & Llama    & 0.0651 & 0.0443 & 0.0374 & 0.0980 & 0.0549 & 0.0418 \\
  & SFR      & 0.0657 & \textbf{0.0464} & \textbf{0.0401} & \textbf{0.0987} & \textbf{0.0571} & \textbf{0.0445} \\
  & Qwen     & \textbf{0.0672} & 0.0456 & 0.0385 & 0.0970 & 0.0552 & 0.0425 \\
\midrule

\multirow{4}{*}{clothing}
  & w/o item pref & 0.0207 & 0.0149 & 0.0130 & 0.0298 & 0.0178 & 0.0142 \\
  & Llama    & 0.0219 & 0.0157 & 0.0137 & \textbf{0.0308} & 0.0186 & 0.0149 \\
  & SFR      & \textbf{0.0227} & \textbf{0.0164} & \textbf{0.0143} & \textbf{0.0308} & \textbf{0.0190} & \textbf{0.0154} \\
  & Qwen     & 0.0210 & 0.0156 & 0.0138 & 0.0299 & 0.0184 & 0.0149 \\
\midrule

\multirow{4}{*}{sport}
  & w/o item pref & 0.0365 & 0.0244 & 0.0204 & 0.0543 & 0.0301 & 0.0227 \\
  & Llama    & 0.0363 & 0.0258 & 0.0224 & 0.0530 & 0.0312 & 0.0246 \\
  & SFR      & 0.0369 & \textbf{0.0264} & \textbf{0.0230} & \textbf{0.0544} & \textbf{0.0320} & \textbf{0.0253} \\
  & Qwen     & \textbf{0.0377} & \textbf{0.0264} & 0.0227 & 0.0534 & 0.0314 & 0.0247 \\
\bottomrule
\end{tabular}
\label{appedix:backbone_comparison}
\end{table*}

\section{Detailed Ablation Study Results}
We conduct extensive ablation study on the top-K strategies and different backbones of PrefRAG cross different datasets. The detailed results are shown in Table~\ref{appendix:top_k_comparison} and Table~\ref{appedix:backbone_comparison}.

\begin{figure*}[t]
\small
\centerline{\includegraphics[width=0.9\textwidth]{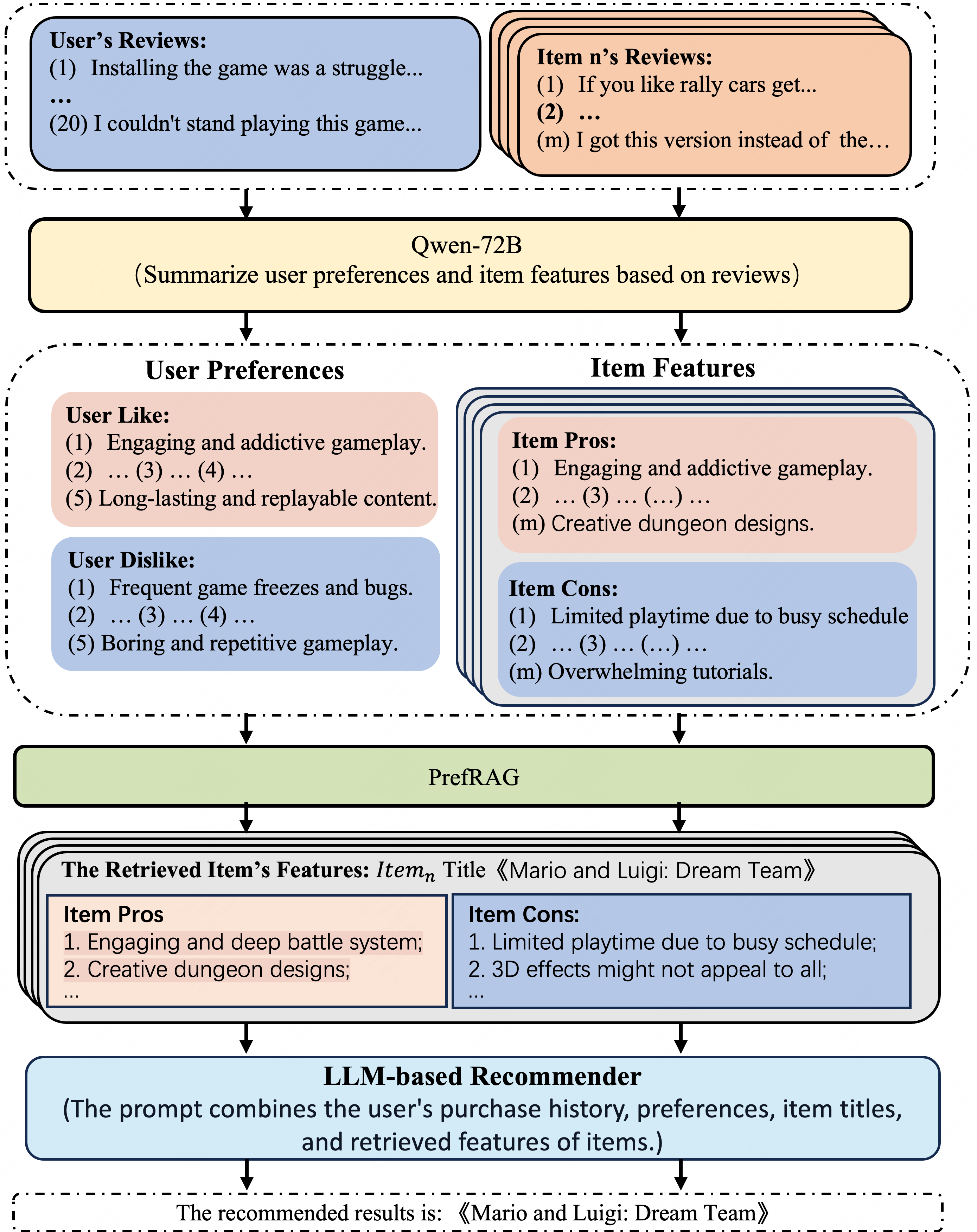}}
\caption{The detailed case study.}
\label{appendix:case study all}
\end{figure*}

\section{Baseline Details}
\label{app:baseline}
To evaluate the effectiveness of our proposed RevBrowse method, we compare it against a diverse set of strong baselines, as detailed below:

\begin{itemize}[leftmargin=1em,labelsep=0.5em]

    \item \textbf{LightGCN}~\cite{DBLP:conf/sigir/0001DWLZ020}: A self-attentive sequential recommendation model that uses Transformer-based attention to capture user behavior patterns from past item interactions.

    \item \textbf{BERT4REC}~\cite{DBLP:conf/cikm/SunLWPLOJ19}: A bidirectional Transformer-based sequential recommendation model that leverages the Cloze task to learn item representations from both past and future user interactions.
    
    \item \textbf{SASRec}~\cite{DBLP:conf/icdm/KangM18}: A self-attention-based sequential recommendation model that adaptively captures user behavior by weighing past interactions to predict next-item preferences, balancing the efficiency of Markov Chains with the expressiveness of RNNs.
    
    \item \textbf{RGCL}~\cite{DBLP:conf/sigir/ShuaiZWSHWL22}: A review-aware graph contrastive learning framework that enhances user-item representation and interaction modeling by leveraging feature-enhanced review signals and dual self-supervised contrastive tasks.
    \item \textbf{DeepCoNN}~\cite{DBLP:conf/wsdm/ZhengNY17}: A deep cooperative neural network model that jointly learns user preferences and item characteristics from review text using parallel CNNs with a shared layer for rating prediction.
    \item \textbf{NARRE}~\cite{DBLP:conf/www/ChenZLM18}: A neural attention-based regression model that jointly predicts user ratings and identifies review-level explanations by weighting reviews based on their usefulness to improve both recommendation accuracy and interpretability.
    \item \textbf{LRU}~\cite{DBLP:journals/corr/abs-2310-02367}: A linear recurrent model for sequential recommendation that combines matrix-diagonalized recurrence and recursive parallelization to achieve fast incremental inference, efficient training, and competitive accuracy across diverse datasets.

    
    \item \textbf{ZS-Ranker}~\cite{DBLP:conf/sigir/Kim0CKCY025}: A prompt-based zero-shot re-ranking model that leverages a large language model to re-rank candidate items based on sequential user interaction histories.
    
    \item \textbf{LlamaRec}~\cite{DBLP:journals/corr/abs-2311-02089}: A re-ranking method that focuses on candidate items, employing LLMs to model and refine item rankings.
    
    \item \textbf{Exp3rt}~\cite{DBLP:journals/corr/abs-2408-06276}: A recommendation model based on user reviews. It summarizes both user profiles and item attributes from review texts to enable the LLM to learn personalized representations for re-ranking.
\end{itemize}

\section{Detailed Case Study}
The detailed illustration of RevBrowse on the typical case is shown in Fig.~\ref{appendix:case study all}.







\section{User Preference and Item Features Extraction Case}
\begin{figure*}[t]
\small
\centerline{\includegraphics[width=0.48\textwidth]{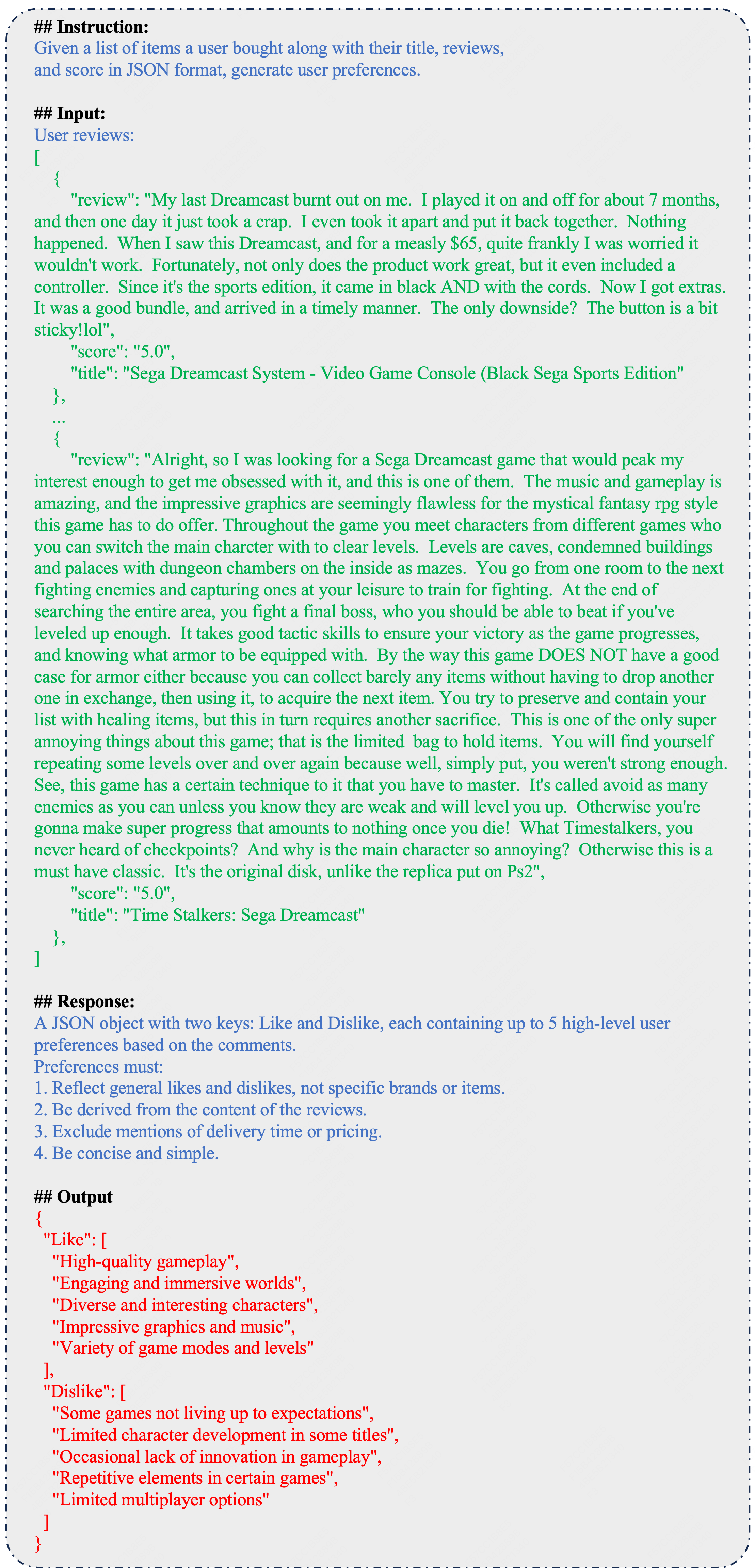}}
\caption{User Preference Extraction Case.}
\label{user_preference_extract}
\end{figure*}

\begin{figure*}[t]
\small
\centerline{\includegraphics[width=0.48\textwidth]{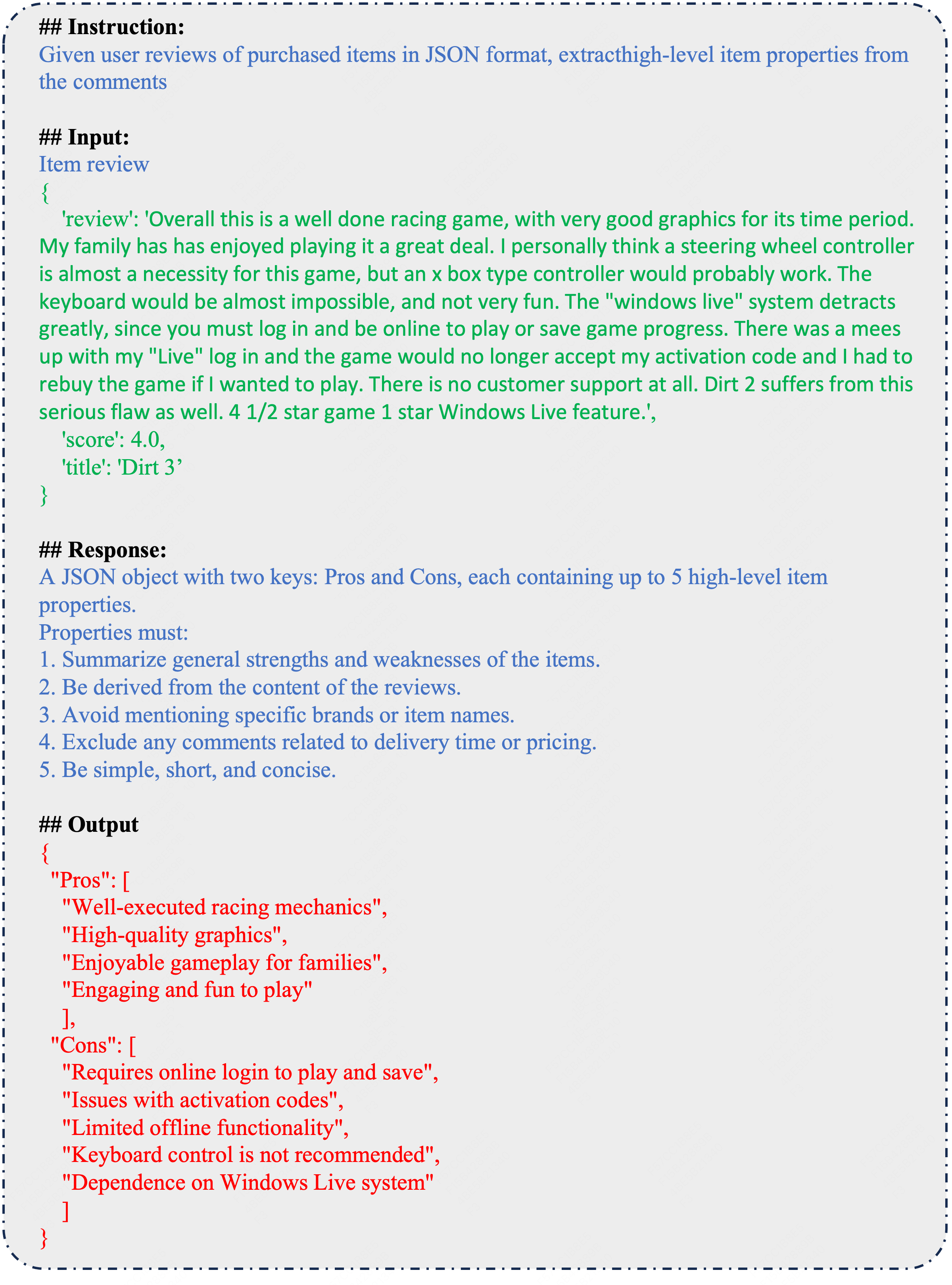}}
\caption{Item Feature Extraction Case.}
\label{item_feature_extract}
\end{figure*}

The input and output example during user preference and item feature extraction is illustrated in Fig.~\ref{user_preference_extract}, and Fig.~\ref{item_feature_extract}.

\end{document}